\documentclass{article}

     \PassOptionsToPackage{numbers, compress}{natbib}

     \usepackage[preprint]{neurips_2021}

\usepackage[utf8]{inputenc} 
\usepackage[T1]{fontenc}    
\usepackage{hyperref}       
\usepackage{url}            
\usepackage{booktabs}       
\usepackage{amsfonts,amssymb}       
\usepackage{nicefrac}       
\usepackage{microtype}      
\usepackage{xcolor}         
\usepackage{pifont}
\usepackage{multirow}
\usepackage{cmath}
\usepackage{graphicx}
\makeatletter 
\newcommand\figcaption{\def\@captype{figure}\caption} 
\newcommand\tabcaption{\def\@captype{table}\caption} 
\title{Tencent Text-Video Retrieval: Hierarchical Cross-Modal Interactions with Multi-Level Representations}

\author{Jie Jiang, Shaobo Min, Weijie Kong, Dihong Gong, Hongfa Wang, Zhifeng Li\thanks{Corresponding authors.}, Wei Liu\footnotemark[1]\\
Tencent Data Platform, Shenzhen, Guangdong, China\\
{\tt\small \{zeus, bobmin, jacobkong, dihonggong, hongfawang, michaelzfli\}@tencent.com}, \\
{\tt\small wl2223@columbia.edu} 
}

\begin{document}
	
	\maketitle
	
\begin{abstract}
	Text-Video Retrieval plays an important role in multi-modal understanding and has attracted increasing attention in recent years.
	Most existing methods focus on constructing contrastive pairs between whole videos and complete caption sentences, while overlooking fine-grained cross-modal relationships,~\emph{e.g.,} clip-phrase or frame-word.
	In this paper, we propose a novel method, named Hierarchical Cross-Modal Interaction (HCMI), to explore multi-level cross-modal relationships among video-sentence, clip-phrase, and frame-word for text-video retrieval.
	Considering intrinsic semantic frame relations, HCMI performs self-attention to explore frame-level correlations and adaptively cluster correlated frames into clip-level and video-level representations.
	In this way, HCMI constructs multi-level video representations for frame-clip-video granularities to capture fine-grained video content, and multi-level text representations at word-phrase-sentence granularities for the text modality.
	With multi-level representations for video and text, hierarchical contrastive learning is designed to explore fine-grained cross-modal relationships,~\emph{i.e.,} frame-word, clip-phrase, and video-sentence, which enables HCMI to achieve a comprehensive semantic comparison between video and text modalities.
	Further boosted by adaptive label denoising and marginal sample enhancement, HCMI achieves new state-of-the-art results on various benchmarks, \emph{e.g.,} Rank@1 of 55.0\%, 58.2\%, 29.7\%, 52.1\%, and 57.3\% on MSR-VTT, MSVD, LSMDC, DiDemo, and ActivityNet, respectively.

\end{abstract}

\section{Introduction}
Text-Video Retrieval (TVR) \cite{luo2021clip4clip,cheng2021improving,gao2021clip2tv,lei2021less,wang2022disentangled,bain2021frozen,xu2021videoclip} has made significant progress in recent years. 
Given a piece of sentence, TVR aims to search a video that is semantically relevant to the targeted sentence from a video database and vice versa.
Compared to other cross-modal video tasks, it is much easier for TVR to obtain paired text-video data from the Internet, such as movies with corresponding captions or YouYube videos with titles.
Thus, based on massive text-video pair data, TVR becomes an important proxy task to understand video contents.
However, a natural semantic gap between two modalities, ~\emph{i.e.,} video and text, raises a great challenge, which hinders industrial-level applications of TVR.
To this end, recent methods target to distill cross-modal knowledge from large-scale pre-training experts \cite{li2020oscar,huang2020pixel,li2021align,rouditchenko2020avlnet} and leverage cross-modal contrastive learning to explore both intra-modal representation and cross-modal interaction.

\begin{figure*}[h]
    \centering
    \includegraphics[width=1\columnwidth]{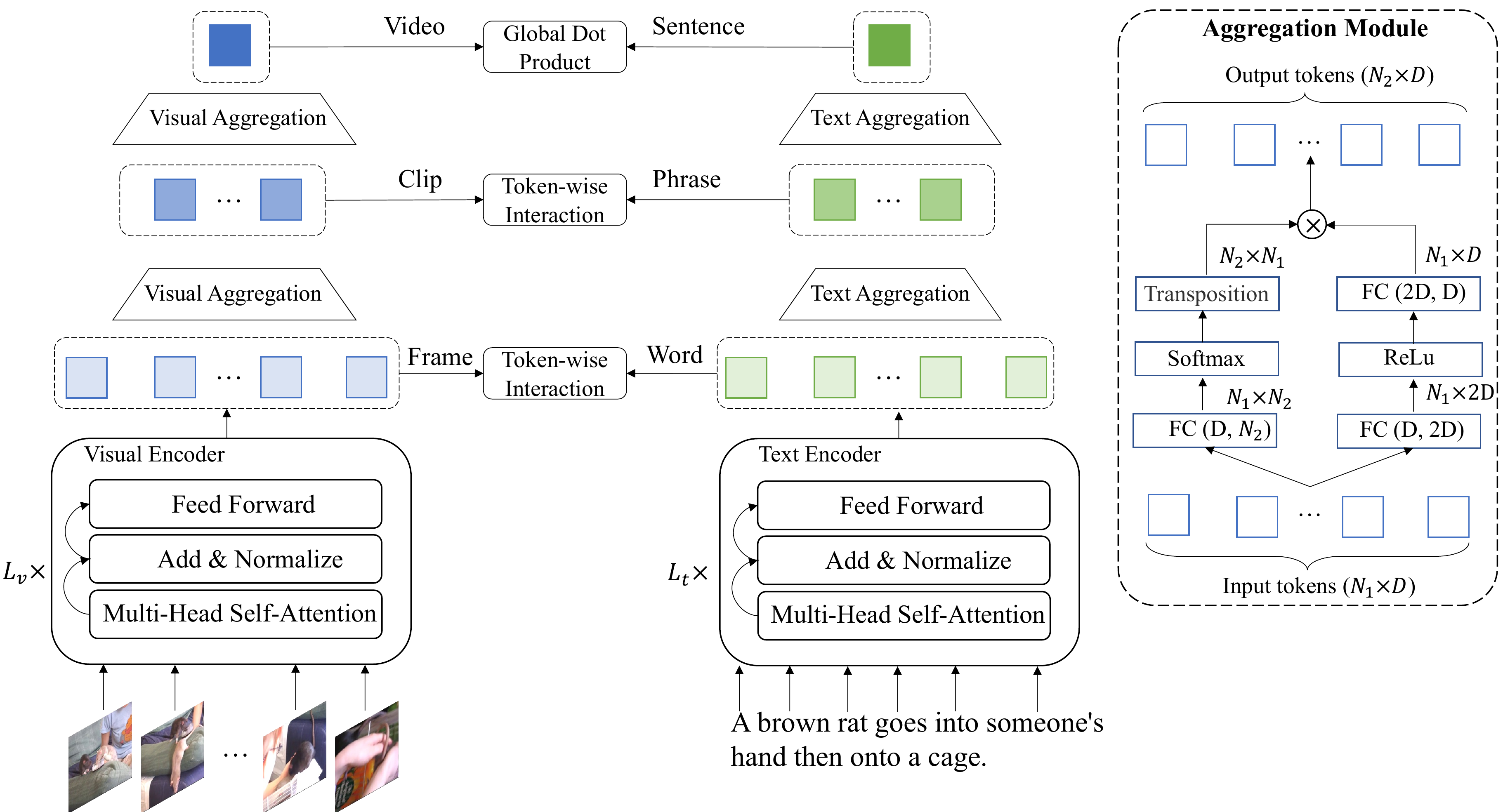}
    \caption{HCMI measures the semantic similarity between a video and a sentence by considering multi-level interactions at frame-word, clip-phrase, and video-sentence granularities, simultaneously. Both visual and text aggregation modules adopt the same architecture, which is given in the right column. FC indicates the fully connected layer. 
    The visual and text encoders adopt standard CLIP architectures in \cite{radford2021learning},~\emph{e.g.,} $L_v=L_t=12$ for CLIP with ViT-B/16.}
    \label{fig:hsi}
\end{figure*}

A pioneering vision-language pre-training work is the Contrastive Language-Image Pretraining (CLIP) \cite{radford2021learning}, which collects 400 million image-text pairs to learn general multi-modal knowledge.
Based on CLIP, recent methods \cite{luo2021clip4clip,gao2021clip2tv,wang2022disentangled,cheng2021improving} aim to transfer the well-pretrained image-text knowledge from text-image to text-video.
For example, CLIP4Clip \cite{luo2021clip4clip} develops temporal ensemble modules to aggregate sequential CLIP frame features into a global one.
Besides, massive matched and unmatched text-video pairs are constructed to learn video representations and text embeddings via cross-modal contrastive learning. 
By fine-tuning on text-video data, CLIP4Clip achieves a great success in TVR.
Different from learning global video representations and sentence embeddings, DCR \cite{wang2022disentangled} proves that the token-wise interaction between frames and words reveals more fine-grained cross-modal knowledge.
Specifically, DCR constructs a similarity matrix between different frame representations and word embeddings, and then infers a comprehensive text-video matching score by considering dense frame-word correlations.
However, these methods only consider single cross-modal interaction from either the video-sentence level or the frame-word level, which results in biased retrieval.
In the human sense, we recognize a video-text pair by simultaneously analyzing video-sentence, clip-phrase, and frame-word interactions, due to the intrinsic hierarchical semantic structure in video and text data, as shown in Figure~\ref{fig:hsi}.

In this paper, we propose a novel method, named Hierarchical Cross-Modal Interaction (HCMI), which hierarchically explores video-sentence, clip-phrase, and frame-word interactions to understand text-video contents comprehensively.
To utilize large-scale pre-training knowledge, HCMI leverages CLIP as initial visual and text encoders, similar to \cite{luo2021clip4clip,wang2022disentangled}.
To explore fine-grained cross-modal interactions, HCMI first constructs hierarchical visual representations and text embeddings at respective frame-clip-video and word-phrase-sentence granularities, as shown in Figure~\ref{fig:hsi}. 
Taking the visual modality as an example, HCMI performs self-attention to gather semantic-correlated frames into several clip representations, which are further fused into a global video representation.
Similar to the video modality, a sentence also has multi-level representations,~\emph{e.g.,} consisting of words and phrases, and can be represented in a word-phrase-sentence manner.
Thus, based on hierarchical video and text representations, HCMI leverages cross-modal contrastive learning to learn inter-modal relationships at frame-word, clip-phrase, and video-sentence granularities, respectively, which achieves a more comprehensive cross-modal comparison than the previous methods.

Besides, we argue that it is unreasonable to regard all videos as individual categories and repel text embeddings from all other video representations with different sample IDs, which is, however, widely used in cross-modal contrastive learning.
For example, the text of 'a man is running' should not be a negative pair of other videos containing 'man running' content.
Thus, HCMI adopts an adaptive label denoising strategy to discover potential positive text-video pairs even with different sample IDs to avoid confusing updating, and proposes a marginal sample enhancement strategy to improve feature discrimination.

Consequently, HCMI obtains new state-of-the-art performance on various benchmarks.
Our contributions are three-fold: 
\begin{enumerate}
\item we propose a novel method, dubbed Hierarchical Cross-Modal Interaction (HCMI), which explores multi-level cross-modal interactions at video-sentence, clip-phrase, and frame-word granularities to understand text-video contents comprehensively;
\item we design adaptive label denoising and marginal sample enhancement strategies to discover potential positive pairs and enlarge hard sample margins, which can avoid noisy gradients and improve feature discrimination; 
\item HCMI obtains new state-of-the-art Rank@1 retrieval results of 55.0\%, 58.2\%, 29.7\%, 52.1\%, 57.3\% on MSR-VTT, MSVD, LSMDC, DiDemo, and ActivityNet, respectively.
\end{enumerate}

This paper is organized as follows.
Section II reviews recent related works on text-video retrieval tasks.
Section III introduces our proposed Hierarchical Cross-Modal Interaction method, as well as adaptive label denoising and marginal sample enhancement.
Section IV gives experimental results and a detailed analysis, followed by the conclusion part in section V.

\section{Related Work}
Previous methods for video understanding focus on designing 3D convolution kernels to capture spatio-temporal information \cite{tran2015learning,xie2018rethinking,feichtenhofer2019slowfast}.
Recently, Vision Transformer (ViT) \cite{dosovitskiy2020image} has shown a great potential in many vision tasks, especially when massive supervision is available.
Thus, many works investigate transformer-based video encoders for video content understanding.

A recent emerging topic is how to explore semantic supervision from large-scale unlabeled data \cite{radford2021learning, miech2020end, radford2021learning}.
In the image-text domain, knowledge transfer and multi-level alignment are two common ways for supervision.
For example, CMR \cite{zhen2020deep} transfers valuable knowledge from existing annotated data to new data via a joint learning paradigm.
DSCMR \cite{zhen2019deep} minimizes the discrimination loss in both the label space and the common representation space to supervise the model, which obtains promising results.
Different from manual labels for video tasks, it is much more convenient to collect large-scale video and text pairs collected from the Internet.
For example, Howto100M \cite{miech2019howto100m} contains millions of instructional videos, and WebVid-2.5M \cite{bain2021frozen} collects 2.5M video-text paris from the web.
Based on massive video-text pairs, plenty of pre-trained model-based methods \cite{luo2020univl, gabeur2020multi,patrick2020support,zhu2020actbert,amrani2021noise,li2020hero,miech2020end,dzabraev2021mdmmt} have dominated the video-text retrieval leaderboard. In general, these methods can be roughly divided into two categories: video-sentence-interaction-based and frame-word-interaction-based methods. 

The video-sentence-interaction-based methods \cite{luo2021clip4clip,liu2019use,portillo2021straightforward,bain2021frozen,croitoru2021teachtext,gabeur2020multi,liu2021hit,liu2019use,khattab2020colbert,kim2021vilt}, whose retrieval process is extremely concise and efficient, employ the separate text and video embedding extractors to map the texts and videos into a common feature space, and then directly conduct the retrieval task based on cosine similarities between their feature representations. For example, ClipBERT \cite{lei2021less} proposes an end-to-end approach with visual and text encoders pre-trained on the image-language dataset and leverages sparse sampling to alleviate the training burden.
FROZEN \cite{bain2021frozen} treats an image as a single-frame video and designs a curriculum learning schedule to train the model on both image and video datasets. Different from FROZEN pre-training a new model on video-text retrieval, the previous SOTO method CLIP4clip \cite{luo2021clip4clip} transfers the knowledge from the image-text pre-trained model CLIP \cite{radford2021learning} to solve the video-text retrieval task. Also based on the pre-trained CLIP model, CAMoE \cite{cheng2021improving} proposes a multi-stream Corpus Alignment network with single gate Mixture-of-Experts (CAMoE) and a novel Dual Softmax Loss (DSL) to further improve the retrieval performance.
The video-sentence-interaction-based methods consider the alignment between global video embedding and sentence embedding. 
However, in the human sense, we compare a video and a sentence in multi-level aspects, such as different frames and words, which is ignored by previous video-sentence-interaction-based methods.

The frame-word-interaction-based methods first utilize embedding extractors to transform each text (video) into a sequence of token (frame) embeddings, and then uses an interaction module to capture fine-grained clues between the token and frame embeddings.
Inspired by \cite{dzabraev2021mdmmt}, the pioneering work FILIP \cite{yao2021filip} proposes a fine-grained interactive language-image method that leverages token-wise maximum similarity between visual and textual tokens, ~\emph{e.g.,} patches and words, to guide the image-text contrastive learning. 
Similarly, DRL \cite{wang2022disentangled} proposes a Weighted Token-wise Interaction to explore the fine-grained clues between sentence tokens and video frame embeddings and a Channel DeCorrelation Regularization to reduce feature redundancy from a micro-view.
Since these methods explore fine-grained similarity measurements between visual-text modalities, they achieve better text-video retrieval performance than video-sentence based methods.

However, almost all of the existing methods only consider single cross-modal interaction from either the video-sentence level or the frame-word level. Few of them explore the intrinsic hierarchical semantic structure in the video and text data. Hence, in this paper, we propose a novel method, dubbed Hierarchical Cross-Modal Interaction, which hierarchically explores video-sentence, clip-phrase, and frame-word interactions to understand text-video contents comprehensively.

\section{Hierarchical Cross-Modal Interaction}
\begin{figure*}[h]
    \centering
    \includegraphics[width=1\columnwidth]{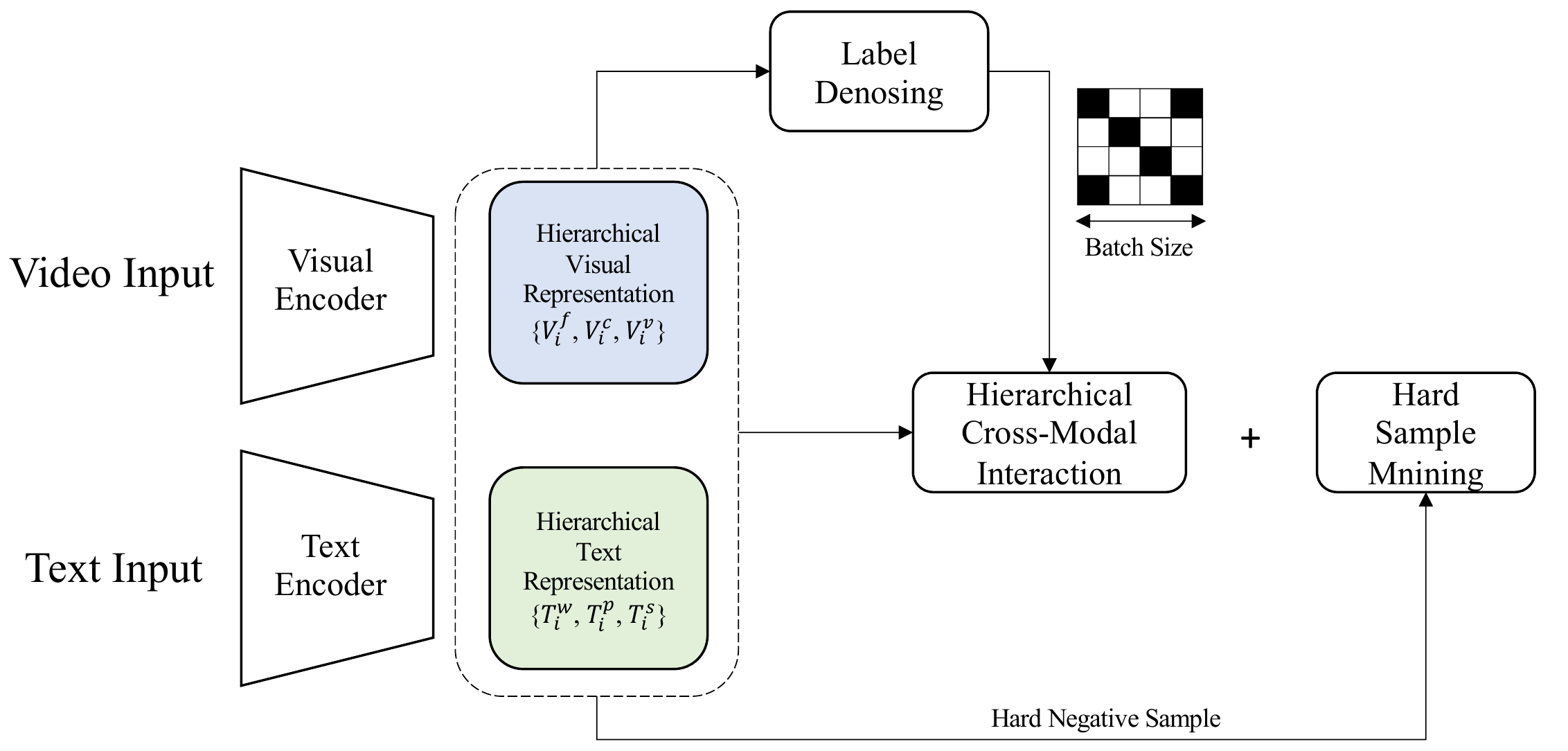}
    \caption{The pipeline of the proposed method. Given hierarchical visual and text features, HCMI first produces text-video pair labels via the adaptive label denoising strategy. Besides, the hard negative samples will be selected according to text-video matching scores. Then, the produced text-video labels and hard negative samples are sent to the hierarchical cross-modal contrastive loss and triplet loss, respectively. 
    }
    \label{fig:pipeline}
\end{figure*}

Given a set of videos $\boldsymbol{v}=\{\boldsymbol{v}_i\}_{i=0}^{N}$ and corresponding captions $\boldsymbol{t}=\{\boldsymbol{t}_i\}_{i=0}^{N}$, our core motivation is to learn a visual encoder $f_v(\cdot)$ and a text encoder $f_t(\cdot)$ that well capture visual and text semantics. 
Here, we denote $V_{i}^f\in R^{N_f\times D}$ as the $N_f$ frame representations extracted from $f_v(\boldsymbol{v}_i)$, and $T_{i}^w\in R^{N_t\times D}$ as the $N_t$ word embeddings extracted from $f_t(\boldsymbol{t}_i)$.
$D$ is the feature dimension.
The overall pipeline of Hierarchical Cross-Modal Interaction (HCMI) is given in Figure~\ref{fig:pipeline}.

\subsection{Multi-Level Representations}
Different from image, video has a hierarchical frame-clip-video structure, which describes the video content from different granularities.
Similarly, the text consists of words, phrases, and global sentences.
In a human sense, we measure the similarity between a video and a text in terms of multiple aspects,~\emph{e.g.,} a frame matches a certain word or a clip matches a certain phrase.
Following this motivation, HCMI designs and learns multi-level video and text representations to compute cross-modal similarity from multiple aspects.

It is noted that $V_{i}^f$ and $T_{i}^w$ contain frame-level and word-level information, which depict fine-grained contents of video and text, respectively.
To further extract features that capture temporal visual information and long-term word dependence, HCMI leverages self-attention to aggregate semantically related frames into clip representations and related words into phrase embeddings, automatically.
Taking visual modality as an example, the aggregation function $g_v(\cdot)$ can be defined as:
\begin{eqnarray}\label{eq:g1}
V_{i}^c = g_v(V_{i}^f)=softmax(V_{i}^fW)^{T} h(V_{i}^f) ~~~~ W\in R^{D\times N_c},
\end{eqnarray}
where $softmax(V_{i}^fW)^{T}$ projects $V_{i}^f\in R^{N_f\times D}$ into normalized frame weights, which has the dimension of $R^{N_f\times N_c}$.
Notably, $N_f$ and $N_c$ are frame number and clip number, respectively.
$h(\cdot)$ is a two-layers FC-ReLU layer with channel changes $D-2D-D$.
Thus, $g_{v}(\cdot)$ aggregates $N_f$ frame representations of $V_i^{f}$ into $N_c$ clip representations of $V_{i}^c \in R^{N_c\times D}$, where $N_c<N_f$.
Denote $V_{i}^c=\{\boldsymbol{v}_{i,1}^c\cdots \boldsymbol{v}_{i,N_c}^c\}$ as a set of $N_c$ clips. $\boldsymbol{v}_{i,j}^c$ actually aggregates several semantically related frame representations into a single one, thereby containing clip information.
Also, the text aggregation function $g_t(\cdot)$ can be given as:
\begin{eqnarray}\label{eq:g2}
T_{i}^p = g_t(T_{i}^w)=softmax(T_{i}^wW)^{T} h(T_{i}^w) ~~~~ W\in R^{D\times N_w},
\end{eqnarray}
where $T_{i}^p\in R^{N_p\times D}$ contains phrase information.
Instead of manually designing clip and phrase information, $V_{i}^c$ and $T_{i}^p$ are produced automatically by aggregating semantically related frames and words.

Similar to $g_v(\cdot)$ and $g_t(\cdot)$, $V_{i}^c$ and $T_{i}^p$ can be further aggregated into the video-level representation $V_i^{v}\in R^{1\times D}$ and the sentence-level embedding $T_i^{s}\in R^{1\times D}$.
Based on $\{V_i^{f}, V_i^{c}, V_i^{v}\}$ and $\{T_i^{w}, T_i^{p}, T_i^{s}\}$, HCMI describes a video and a sentence at frame-clip-video and word-phrase-sentence granularities, respectively.
Next, we propose hierarchical contrastive learning for different granularities.

For $\{V_i^{f},T_i^{w}\}$, the token-wise interaction function is given by:
\begin{eqnarray}\label{eq:f-w}
\begin{aligned}
\mathcal{L}_{f-w} =& -\frac{1}{2N}\sum_{i}^{N}(log\frac{exp(TI(V_i^{f},T_i^{w}))}{\sum_{j}^{N}exp(TI(V_i^{f},T_j^{w}))}
+log\frac{exp(TI(T_i^{w},V_i^{f}))}{\sum_{j}^{N}exp(TI(T_i^{w},V_j^{f}))}), 
\end{aligned}
\end{eqnarray}

\begin{eqnarray}\label{eq:f-w1}
\begin{aligned}
TI(V_i^{f},T_i^{w}) =& \big(\frac{1}{2N_t}\sum^{N_t}_{n=1}max_{m=1}^{N_v}<T_{i,n}^w,V_{i,m}^{f}> 
+\frac{1}{N_v}\sum^{N_v}_{n=1}max_{m=1}^{N_t}<V_{i,n}^f,T_{i,m}^{w}>\big),
\end{aligned}
\end{eqnarray}
where $<\cdot,\cdot>$ is the dot product function.
$TI(V_i^{f},T_i^{w})$ first calculates a pair-wise similarity matrix between frames and words and then aggregates all token-wise similarities into an overall score.
$\mathcal{L}_{f-w}$ is a symmetric cross-modal contrastive loss that measures the cross-modal similarity between a set of frames and words.

For $\{V_i^{c},T_i^{p}\}$, $\mathcal{L}_{c-p}$ is given by:
\begin{eqnarray}\label{eq:c-p}
\begin{aligned}
\mathcal{L}_{c-p} =& -\frac{1}{2N}\sum_{i}^{N}(log\frac{exp(TI(V_i^{c},T_i^{p}))}{\sum_{j}^{N}exp(TI(V_i^{c},T_j^{p}))}
+log\frac{exp(TI(T_i^{p},V_i^{c}))}{\sum_{j}^{N}exp(TI(T_i^{p},V_j^{c}))}).
\end{aligned}
\end{eqnarray}
Similar to $\mathcal{L}_{f-w}$, $\mathcal{L}_{c-p}$ uses $TI(\cdot,\cdot)$ to measure the cross-modal similarity between clips and phrases.

For $\{V_i^{v},T_i^{s}\}$, $\mathcal{L}_{v-s}$ is given by:
\begin{eqnarray}\label{eq:v-s}
\begin{aligned}
\mathcal{L}_{v-s} =& -\frac{1}{2N}\sum_{i}^{N}(log\frac{exp(<V_i^{v},T_i^{s}>)}{\sum_{j}^{N}exp(<V_i^{v},T_j^{s}>)} 
+log\frac{exp(<T_i^{s},V_i^{v}>)}{\sum_{j}^{N}exp(<T_i^{s},V_j^{v}>)}).
\end{aligned}
\end{eqnarray}
Here, $\mathcal{L}_{v-s}$ uses cosine similarity to measure the cross-modal similarity between video and sentence representations.

Finally, the loss function for hierarchical cross-modal interaction is:
\begin{eqnarray}\label{eq:L}
\mathcal{L}_{hci} = \mathcal{L}_{f-w} + \alpha\mathcal{L}_{c-p} + \beta\mathcal{L}_{v-s},
\end{eqnarray}
where $\alpha$ and $\beta$ balance different terms.

\subsection{Adaptive Label Denoising}
Furthermore, most existing methods simply treat all video-text pairs with different sample IDs as negative pairs.
For example, a text has only one positive video that is from the same data pair, and all the videos from other data pairs are treated as negative datapoints.
However, in fact, one text usually has other similar video descriptions in the dataset, so when taking such datapoints as the negative ones, it will harm the retrieval performance of the model.

To tackle the above issue, we design a novel adaptive label denoising scheme to discover such potential similar datapoints. 
Specifically, for each video in a data batch, we target to discover other videos with similar content, which cannot be treated as negative pairs.
For a video $\boldsymbol{v}_i$, we obtain two views $\boldsymbol{v}_i^1$ and $\boldsymbol{v}_i^2$ by randomly sampling its frames twice. 
We define two videos to be similar, if $\boldsymbol{v}_i^1$ and $\boldsymbol{v}_j^1$ are more similar than $\boldsymbol{v}_i^1$ and $\boldsymbol{v}_i^2$.
In other words, the feature embedding of these two views,~\emph{i.e.,} $V_i^1$ and $V_i^2$, should be similar, as they describe the same video content.
Similarly, supposing there are two views $V_j^1$ and $V_j^2$ for the video $\boldsymbol{v}_j$, the video $\boldsymbol{v}_j$ can be treated similar to $\boldsymbol{v}_i$, when the following inequality holds:
\begin{equation}
	\begin{aligned}		
	cos(V_i^{1}, V_j^{1}) \ge cos(V_i^{1}, V_i^{2}),
	\end{aligned}
	\label{eq:similar}
\end{equation} 
where $cos(\cdot,\cdot)$ denotes the cosine similarity between the inputs.
Here, $V_i^{k}$ represents the video level representation of the $k^{th}$ view of video $\boldsymbol{v}_i$.
Based on Eq.~\eqref{eq:similar}, we can obtain pair-wise similarities in a data batch. For example, for the $i$-th sample, we collect all the samples that meet the conditions of Eq.~\eqref{eq:similar}, and these samples make up a subset $N^{+}_{i}$. Notably, the samples in $N^{+}_{i}$ are regarded as similar videos with $i$-th sample, which cannot be treated as negative samples.
 
 In a data batch, when videos $\boldsymbol{v}_i$ and $\boldsymbol{v}_j$ are defined similar through Eq. (\ref{eq:similar}), the video-text pairs ($\boldsymbol{v}_i$,  $\boldsymbol{t}_i$) and ($\boldsymbol{v}_j$,  $\boldsymbol{t}_j$) are considered as positive.
 Thus, Eqs. (\ref{eq:f-w}), (\ref{eq:c-p}) and (\ref{eq:v-s}) can be reformulated as follows:
 \begin{eqnarray}\label{eq:f-w1}
 \begin{aligned}
\mathcal{L}_{f-w} = &-\sum_{i}^{N}(\sum_{k\in N_i^{+}}log\frac{exp(TI(V_i^{f},T_k^{w}))}{\sum_{j\in N_i^{-}\cup \{k\}}exp(TI(V_i^{f},T_j^{w}))}\\
&+\sum_{k\in N_i^{+}}log\frac{exp(TI(T_i^{w},V_k^{f}))}{\sum_{j\in N_i^{-}\cup \{k\}}exp(TI(T_i^{w},V_j^{f}))})/2N,
\end{aligned}
\end{eqnarray}
 \begin{eqnarray}\label{eq:c-p1}
 \begin{aligned}		
\mathcal{L}_{c-p} = &-\sum_{i}^{N}(\sum_{k\in N_i^{+}}log\frac{exp(TI(V_i^{c},T_k^{p}))}{\sum_{j\in N_i^{-}\cup \{k\}}exp(TI(V_i^{c},T_j^{p}))}\\
&+\sum_{k\in N_i^{+}}log\frac{exp(TI(T_i^{p},V_k^{c}))}{\sum_{j\in N_i^{-}\cup \{k\}}exp(TI(T_i^{p},V_j^{c}))})/2N,
\end{aligned}		
\end{eqnarray}
 
 \begin{eqnarray}\label{eq:v-s1}
 \begin{aligned}		
\mathcal{L}_{v-s} = &-\sum_{i}^{N}(\sum_{k\in N_i^{+}}log\frac{exp(<V_i^{v},T_k^{s}>)}{\sum_{j\in N_i^{-}\cup \{k\}}exp(<V_i^{v},T_j^{s}>)} \\
&+\sum_{k\in N_i^{+}}log\frac{exp(<T_i^{s},V_k^{v}>)}{\sum_{j\in N_i^{-}\cup \{k\}}exp(<T_i^{s},V_j^{v}>)})/2N,
\end{aligned}		
\end{eqnarray}
where $N_i^{+}$'s denote indexes of data pairs similar to the data pair ($\boldsymbol{v}_i$, $\boldsymbol{t}_i$); $N_i^{-}$'s denote the indexes of data pairs dissimilar to ($\boldsymbol{v}_i$, $\boldsymbol{t}_i$).

Compared with Eq. (\ref{eq:f-w}), (\ref{eq:c-p}) and (\ref{eq:v-s}), Eq. (\ref{eq:f-w1}), (\ref{eq:c-p1}) and (\ref{eq:v-s1}) remove some false negative video-text pairs from $N_i^{-}$, thereby rendering faster convergence and better retrieval performance.

\begin{table*}[t]
    \centering
    \caption{The ablation study results on MSR-VTT. 'Base' indicates the baseline CLIP4Clip model. 'GDP' is the global dot product interaction, and 'TWI' uses token-wise interaction. 'HCI' is our proposed hierarchical cross-modal loss $\mathcal{L}_{hci}$. 'Denoise' and 'MSE' are the adaptive label denoising and marginal sample enhancement strategies, respectively. 'Dual' is the post-processing of dual softmax.}
    \resizebox{1\columnwidth}{!}{ 
    \begin{tabular}{l|ccc|ccc|ccccc}
	\hline
	\multirow{2}{*}{Base} & \multirow{2}{*}{GDP} & \multirow{2}{*}{TWI} & \multirow{2}{*}{HCI} & \multirow{2}{*}{Denoise} & \multirow{2}{*}{MSE} & \multirow{2}{*}{Dual} & \multicolumn{5}{c}{Text-Video Retrieval}\\ 
    &&&&&&& R@1 & R@5 & R@10 & MdR & MnR\\ 
	\hline
	\hline
	\checkmark&\checkmark&&&&&&44.6&72.0&82.3&2.0&13.6\\
	\checkmark&&\checkmark&&&&&48.3&74.9&83.8&2.0&12.4\\
	\checkmark&&&\checkmark&&&&49.0&76.5&84.3&2.0&11.8\\
	\checkmark&&&\checkmark&\checkmark&&&49.5&76.9&84.2&2.0&12.2\\
	\checkmark&&&\checkmark&\checkmark&\checkmark&&49.7&75.0&83.5&2.0&11.4\\
	\checkmark&&&\checkmark&\checkmark&\checkmark&\checkmark&54.9&79.5&87.0&1.0&10.2\\
	\hline
    \end{tabular}}
    \label{tab:ablation}
\end{table*}

\subsection{Marginal Sample Enhancement}
During optimizing the contrastive loss, it is easy for the model to distinguish negative sample pairs with obviously different contents, which contain limited information.
Thus, the model should focus more on those hard samples with a subtle content difference, and we design a marginal sample enhancement loss to emphasize the distinguishing capacity for hard text-video pair samples.

Concretely, for a video $V_{i}^{v}$, we select the hardest text sample $T_{j}^{s}$ according to $\mathcal{L}_{hci}$ and vice versa.
Then, a triplet loss is applied to $\{V_{i}^{v}, T_{i}^{s}, T_{j}^{s}\}$ and $\{T_{i}^{s}, V_{i}^{v}, V_{k}^{v}\}$ as:
\begin{eqnarray}\label{eq:L_hsm}
 \begin{aligned}	
\mathcal{L}_{hsm} =& (max(<V_{i}^{v}, T_{j}^{s}>-<V_{i}^{v}, T_{i}^{s}> + \theta,0) \\
& + max(<T_{i}^{s}, V_{k}^{v}>-<T_{i}^{s}, V_{i}^{v}> + \theta,0))/2, 
\end{aligned}	
\end{eqnarray}
where $\theta$ is a margin coefficient.
Here, we only use $\{V_{i}^{v}, T_{i}^{s}\}$ at the video-sentence granularity, as the gradients will be back-propagated to $\{V_{i}^{f}, T_{i}^{w}\}$ and $\{V_{i}^{c}, T_{i}^{p}\}$.

\subsection{Overall Objective}
The overall objective function of HCMI is
\begin{eqnarray}\label{eq:L_over}
\mathcal{L} = \mathcal{L}_{hci} + \lambda\mathcal{L}_{hsm}, 
\end{eqnarray}
where $\lambda$ controls the balance across two terms.

\section{Experiments}
Experiments are conducted on five Text-Video Retrieval (TVR) benchmarks, and several ablation studies are given to demonstrate the effectiveness of each component in the proposed Hierarchical Cross-Modal Interaction (HCMI).

\subsection{Datasets and Evaluation Metrics}
HCMI is evaluated on five public benchmarks:

\noindent\textbf{MSR-VTT}\cite{xu2016msr} is the most popular TVR benchmark, which contains 10,000 videos with 20 captions. We report the results on the standard full split.

\noindent\textbf{MSVD}\cite{wu2017deep} contains 1,970 videos with 80,000 captions. We report the standard split.

\noindent\textbf{DiDeMo}\cite{rohrbach2015dataset} contains 10,000 videos with 40,000 sentences. Following \cite{luo2021clip4clip}, all captions are concatenated into a single query for text-video retrieval.

\noindent\textbf{LSMDC}\cite{anne2017localizing} contains 118,081 videos with an equal number of caption sentences from 202 movies. We adopt the standard split for training and testing.

\noindent\textbf{ActivityNet}\cite{caba2015activitynet} contains 20,000 YouTube videos. Following \cite{luo2021clip4clip}, we concatenate all captions of a video as a single query.

Following \cite{luo2021clip4clip}, we report experiments under the standard TVR metrics at rank-K, which evaluate the percentage of query samples for which the right answer is founded in the tok-K retrieved results.
We report the rank-1, rank-5, rank-10, median rank, and mean rank metrics. 
The median rank and mean rank calculate the median and mean rank of all correct results.
Notably, higher rank-1, rank-5, and rank-10 are better, and lower median rank and mean rank are better.

\subsection{Implementation Details}
The basic visual and text encoders adopt the pre-trained weights in CLIP \cite{radford2021learning}, which include ViT-B/16 and ViT-L/14 architectures \cite{dosovitskiy2020image}.
Besides, a 4-layer temporal transformer is added to capture the temporal information on top of the visual encoder.
Similar to CLIP4Clip \cite{luo2021clip4clip}, parameters of the temporal transformer are initialized from the first 4 layers of the text encoder in CLIP.
The frame length $N_v$ and word length $N_w$ are 12 and 32 for MSR-VTT, MSVD, LSMDC, and 64 and 64 for DiDeMo and ActivityNet, respectively.
The network is optimized by Adam with 5 epochs.
The batch size is 128 for ViT-B/16 and 64 for ViT-L/14.
The initial learning rate is $1e-7$ for the clip parameters and $1e-4$ for the non-clip parameters, respectively.
The hyper-parameters are set as $N_c=N_p=6$, $\alpha=0.5$, $\beta=0.1$, and $\theta=\lambda=0.1$, which will be analyzed in the ablation study. 

We use 32 A100 GPUs for the ViT-L/14 backbone and 8 A100 GPUs for the ViT-B/16 backbone under the 128 batch-size setting for experiments. 
Each GPU is equipped with 8 CPU cores and 48G RAM.
The algorithm is implemented via python code.

For the HCMI with SMoE, we explore the best setting for different datasets using 64 A100
GPUs. Here, we utlize the ViT-L/14 as the basic model, which has aroud 450M parameters. Then,
we replace the FFN layers in ViT-L/14 with SMoE layers to expand the model scale. For MSR-VTT,
LSMDC and MSVD with 12 frames and 32 words, we use 64 experts in each SMoE layer, which
produces about 17B parameters of the whole model. For ActivityNet and DiDeMo with 32 frames
and 64 words, we use 8 experts in each SMoE layer to fill up the GPUs. Finally, after experiments,
We find that $k = 2$ gives the best results for MSR-VTT and LSMDC, and $k = 4$ is the best for
MSVD. For ActivityNet and DiDeMo, $k = 1$ gives the best results.

\subsection{Ablation Study}
In this part, we analyze the effect of each component and hyper-parameter in HCMI and visualize some important results.

\begin{figure*}[h]
    \centering
    \includegraphics[width=1.0\columnwidth]{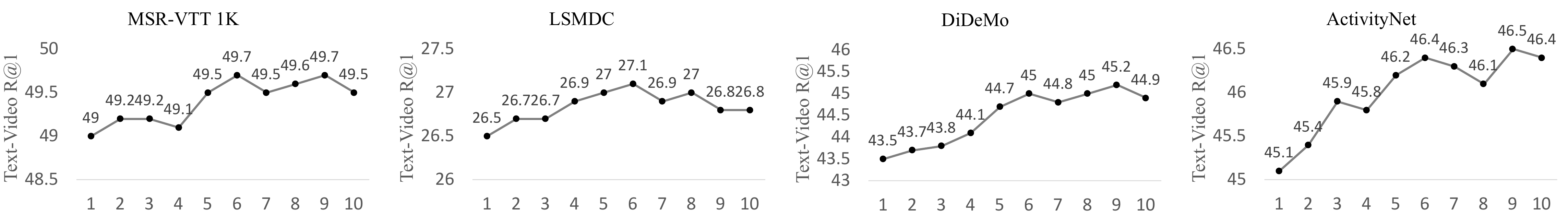}
    \caption{Evaluation of $N_c$ on different datasets.}
    \label{fig:hc_hp}
\end{figure*}

\begin{figure*}[h]
    \centering
    \includegraphics[width=1\columnwidth]{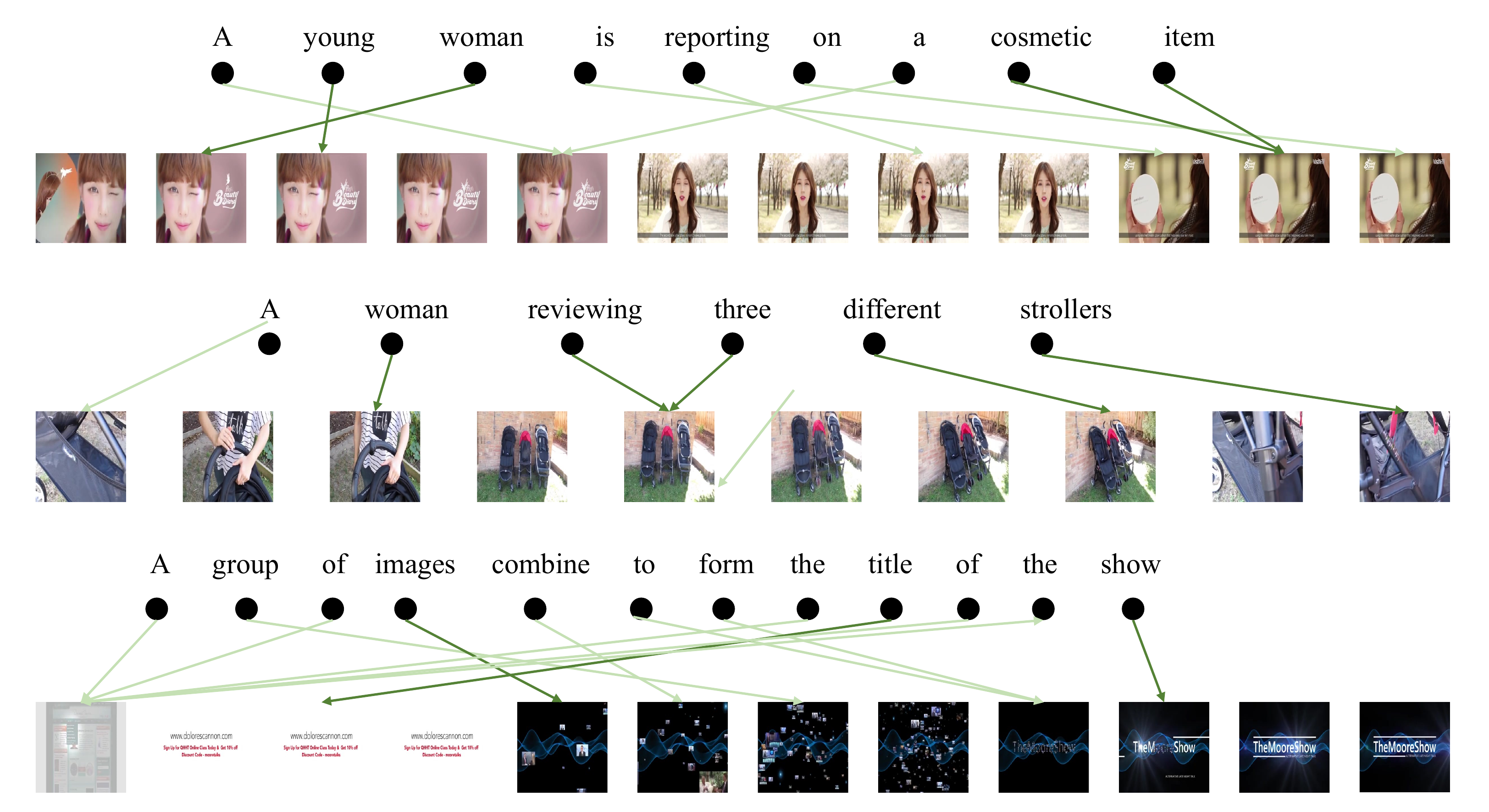}
    \caption{Some examples of fine-grained frame-word matching results in HCMI, \emph{e.g.,} the best matching frame for each word. The deep green line indicates matching with a large similarity.}
    \label{fig:vis}
\end{figure*}

\begin{figure*}[h]
    \centering
    \includegraphics[width=1\columnwidth]{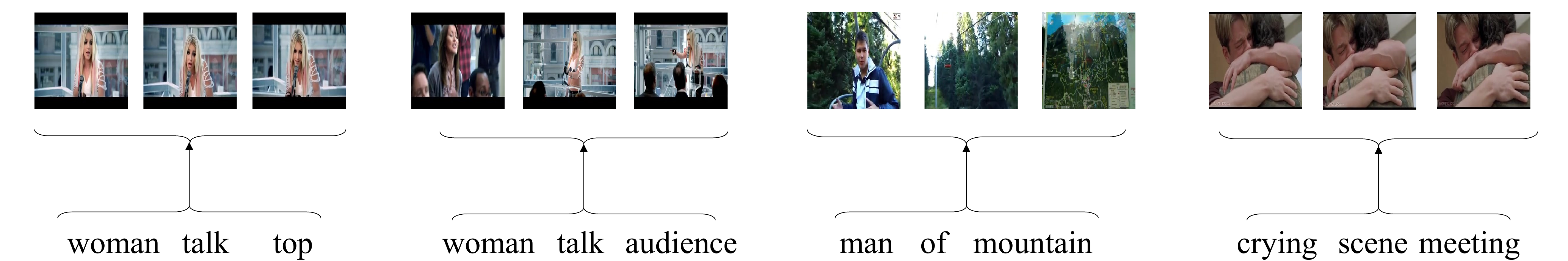}
    \caption{Some examples of auto-aggregated clip-phrase matching results in HCMI. For each clip and phrase, we select top-3 tokens to visualize.}
    \label{fig:vis_cp}
\end{figure*}

\subsubsection{Evaluation on each component}
We conduct experiments on MSR-VTT to evaluate the effect of each module and method in HCMI.
'Base' indicates the baseline CLIP4Clip model with a temporal transformer and ViT-B/16.
'GDP' is the global dot product interaction between video representations and sentence embeddings, which is used in the CLIP4Clip method.
'TWI' uses token-wise interaction in \cite{wang2022disentangled}, which explores a dense relationship between different frames and words.
'HCI' is the multi-level cross-modal interaction in HCMI, which explores frame-word, clip-phase, and video-word relationships, simultaneously.
'Denoise' and 'MSE' indicate adding adaptive label denoising and marginal sample enhancement strategies.
'Dual' means using the dual-softmax proposed in \cite{cheng2021improving}.
Notably, 'Dual' is only used during the inference stage as a post-processing strategy.

\begin{figure*}[h]
    \centering
    \includegraphics[width=1.0\columnwidth]{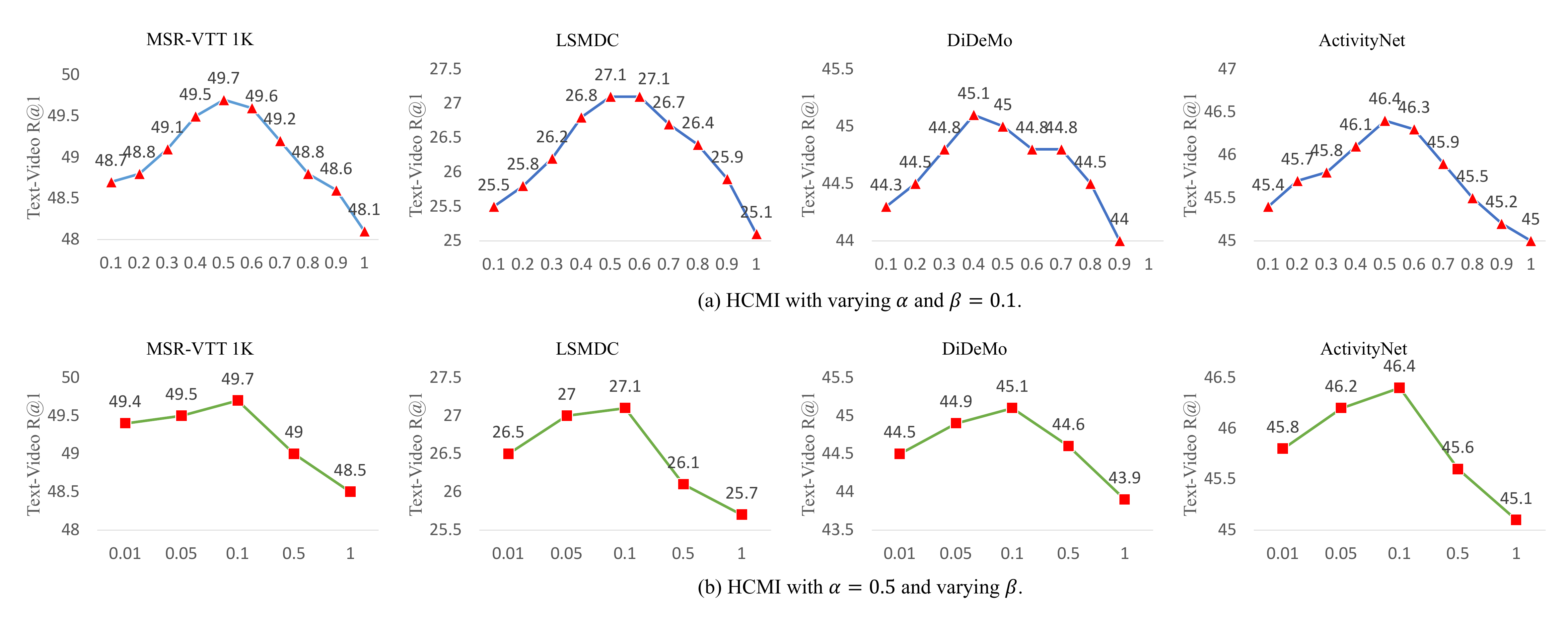}
    \caption{Evaluation of $\alpha$ and $\beta$ on different datasets.}
    \label{fig:ab}
\end{figure*}

\begin{table*}[t]
    \centering
    \caption{Retrieval results on MSR-VTT 1K. * indicates that the method uses post-processing operations, ~\emph{e.g.,} DSL \cite{cheng2021improving} or QB-Norm \cite{bogolin2021cross}.}
    \resizebox{1\columnwidth}{!}{ 
    \begin{tabular}{l|ccccc|ccccc}
	\hline
	\multirow{2}{*}{Methods} & \multicolumn{5}{c|}{Text-Video} & \multicolumn{5}{c}{Video-Text} \\
    &R@1&R@5&R@10&MdR&MnR& R@1 & R@5 & R@10 & MdR & MnR\\ 
	\hline
	\hline
	HERO\cite{li2020hero}&16.8&43.4&57.7&-&&-&-&-&-&-\\
	UniVL\cite{luo2020univl}&21.2&49.6&63.1&6.0&-&-&-&-&-&-\\
	ClipBERT\cite{lei2021less}&22.0&46.8&59.9&6.0&-&-&-&-&-&-\\
	MDMMT\cite{dzabraev2021mdmmt}&26.6&57.1&69.6&4.0&-&27.0&57.5&69.7&3.7&-\\
	SUPPORT\cite{patrick2020support}&27.4&56.3&67.7&3.0&-&26.6&55.1&67.5&3.0&-\\
	FROZEN\cite{bain2021frozen}&31.0&59.5&70.5&3.0&-&-&-&-&-&-\\
	CLIP4Clip\cite{luo2021clip4clip}&44.5&71.4&81.6&2.0&15.3&42.7&70.9&80.6&2.0&-\\
	CLIP2Video\cite{fang2021clip2video}&45.6&72.6&81.7&2.0&14.6&43.5&72.3&82.1&2.0&10.2\\
	CAMoE*\cite{cheng2021improving}&48.8&75.6&85.3&2.0&12.4&50.3&74.6&83.8&2.0&9.9\\
	CLIP2TV*\cite{gao2021clip2tv}&52.9&78.5&86.5&1.0&12.8&54.1&77.4&85.7&1.0&9.0\\
	MDMMT-2\cite{kunitsyn2022mdmmt}&48.8&75.7&84.4&2.0&13.5&-&-&-&-&-\\
	DCR*\cite{wang2022disentangled}&53.3&80.3&87.6&1.0&-&$\boldsymbol{56.2}$&79.9&87.4&1.0&-\\
	\hline
	Ours$_{+B/16}$&49.7&75.0&83.5&2.0&11.4&47.4&76.1&85.2&2.0&8.1\\
	Ours$_{+L/14}$&49.5&74.2&83.9&2.0&12.1&46.8&75.1&84.0&2.0&9.7\\
	Ours$_{+B/16+DSL}$&$\boldsymbol{55.0}$&80.4&86.8&1.0&10.3&55.5&78.4&85.8&1.0&7.7\\
	\hline
    Ours$_{huge}$&$\boldsymbol{62.9}$&84.5&90.8&1.0&9.3&64.8&84.9&91.1&1.0&5.5\\
	\hline
    \end{tabular}}
    \label{tab:msrvtt}
    \vspace{-0.2cm}
\end{table*}

\begin{table*}[t]
    \centering
    \caption{Retrieval results on MSVD. * indicates that the method uses post-processing operations, ~\emph{e.g.,} DSL \cite{cheng2021improving} or QB-Norm \cite{bogolin2021cross}.}
    \resizebox{1\columnwidth}{!}{ 
    \begin{tabular}{l|ccccc|ccccc}
	\hline
	\multirow{2}{*}{Methods} & \multicolumn{5}{c|}{Text-Video}& \multicolumn{5}{c}{Video-Text}\\ 
    &R@1&R@5&R@10&MdR&MnR& R@1 & R@5 & R@10 & MdR & MnR\\ 
	\hline
	\hline
	CE\cite{liu2019use}&19.8&49.0&63.8&6.0&-&-&-&-&-&-\\
	SUPPORT\cite{patrick2020support}&28.4&60.0&72.9&4.0&-&-&-&-&-&-\\
	FROZEN\cite{bain2021frozen}&33.7&64.7&76.3&3.0&-&-&-&-&-&-\\
	CLIP4Clip\cite{luo2021clip4clip}&46.2&76.1&84.6&2.0&10.0&48.4&70.3&77.2&2.0&-\\
	CAMoE*\cite{cheng2021improving}&49.8&79.2&87.0&-&9.4&-&-&-&-&-\\
	MDMMT-2\cite{kunitsyn2022mdmmt}&56.8&83.1&89.2&1.0&8.0&-&-&-&-&-\\
	DCR*\cite{wang2022disentangled}&50.0&81.5&89.5&2.0&-&58.7&92.5&95.6&1.0&-\\
	\hline
	Ours$_{+B/16}$&49.7&79.0&87.7&2.0&8.9&59.9&85.9&90.0&1.0&5.6\\
	Ours$_{+L/14}$&52.7&80.4&88.3&1.0&8.4&68.0&90.5&94.8&1.0&3.9\\
	Ours$_{+L/14+DSL}$&$\boldsymbol{58.2}$&83.5&90.1&1.0&7.8&$\boldsymbol{69.1}$&91.5&95.0&1.0&3.8\\
	\hline
    Ours$_{huge}$&$\boldsymbol{59.0}$&84.0&90.3&1.0&7.6&$\boldsymbol{73.0}$&94.5&96.6&1.0&7.6\\
	\hline
    \end{tabular}}
    \label{tab:msvd}
    \vspace{-0.2cm}
\end{table*}

\begin{figure}[h]
    \centering
    \includegraphics[width=0.5\columnwidth]{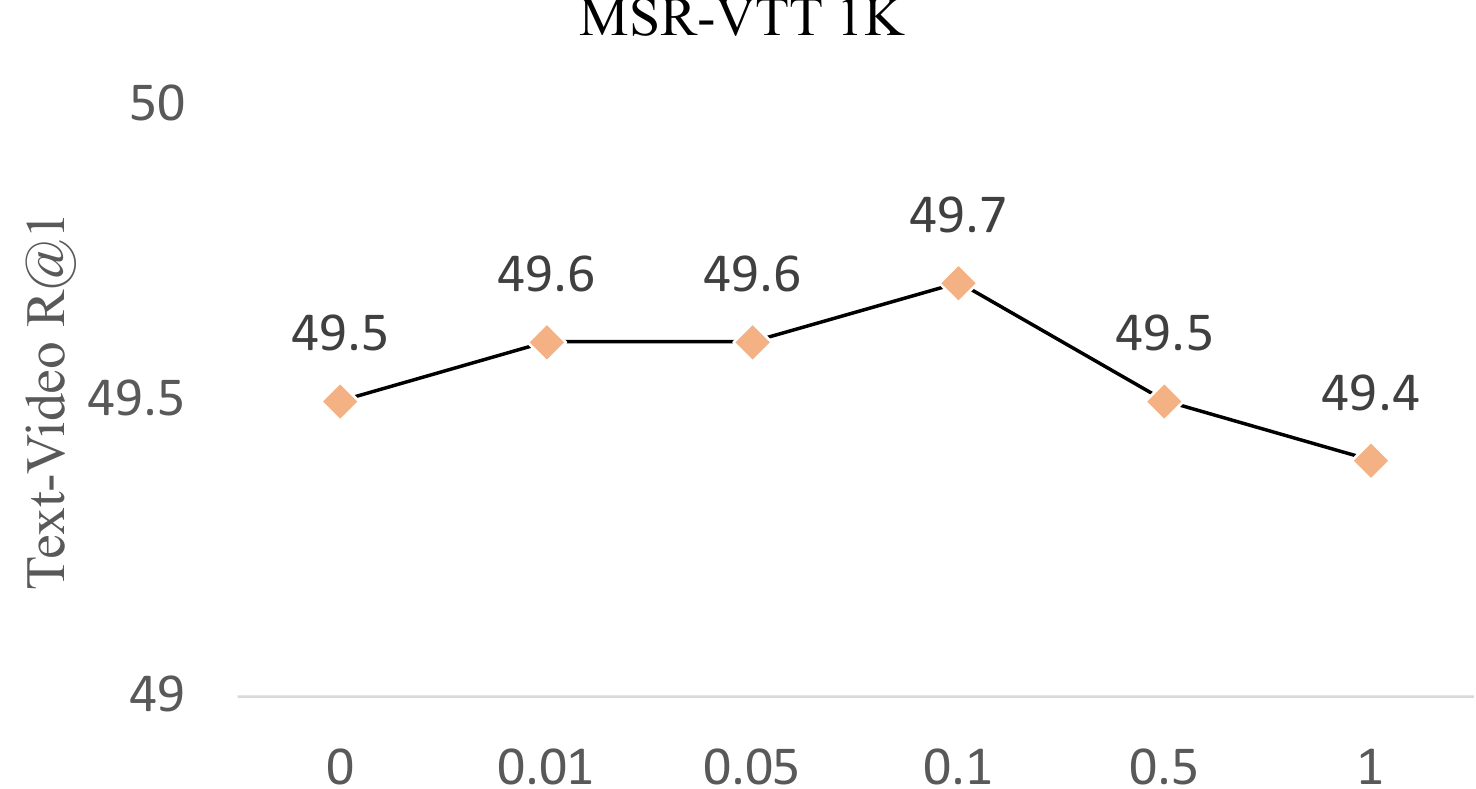}
    \caption{Evaluation of $\lambda$ on MSR-VTT.}
    \label{fig:lambda}
\end{figure}

The results are given in Table~\ref{tab:ablation}.
It can be seen that, compared to the global dot product interaction in 'Base', our hierarchical cross-modal interaction ('HCI') brings a significant gain,~\emph{e.g.,} 4.9\% improvement on R@1.
Compared to the token-wise interaction 'TWI', 'TCI' also shows obvious superiority, due to extra clip-phrase and video-sentence interactions.
This demonstrates that considering multi-level interaction between video and text can significantly improve performance.
Besides, both adaptive label denoising and marginal sample enhancement further bring gain, which shows that HCMI can successfully discover potential position samples in a batch.
Finally, our HCMI is made up of Base+HCI+Denoise+MSE+Dual.

\subsubsection{Evaluation on hyper-parameters}
In this part, we evaluate the effect of each hyper-parameter in HCMI on four benchmarks,~\emph{i.e.,} MSR-VTT, LSMDC, DiDeMo, and ActivityNet.
Here, the post-processing strategy of Dual Softmax is not used.

First, we conduct experiments on different $N_c$ or $N_p$.
Here, we set $N_c = N_p$, which indicates how many clips or phrases HCMI obtains from frames or words.
The results are given in Figure~\ref{fig:hc_hp}.
It can be seen that, when $N_c$ is larger than $6$ on MSR-VTT and LSMDC, the improvement becomes weak.
The reason is that we sample 12 frames for MSR-VTT and LSMDC datasets, so $N_c=6$ is enough to capture the clip information.
However, for DiDeMo and ActivityNet with 64 frames, large $N_c$ gives an obvious gain.
In consideration of model simplicity and complexity, we choose $N_c=6$  for all the datasets to achieve an overall performance on the five datasets.

Then, we evaluate the effect of different $\alpha$ and $\beta$.
In Figure~\ref{fig:ab}, we find that $\alpha=0.5$ and $\beta=0.1$ give the best performance.
An interesting phenomenon is that, when $\alpha>0.5$ or $\beta>0.1$, the performance drops seriously.
The reason is that, the frame-word interaction in Eq.~\eqref{eq:f-w} and Eq.~\eqref{eq:f-w1} also contains the clip-phrase and video-sentence information, as frames and words make up of clip and phrase.
Thus, when using large $\alpha$ and $\beta$, it will lead to unbalanced loss contributions among frame-word, clip-phrase, and video-sentence granularities.
Besides, the automatically aggregated clip representations and phrase embeddings also contain video- and sentence-level information.
Thus, the optimal $\alpha$ is larger than $\beta$, and both $\alpha$ and $\beta$ are sensitive to a large value.

\begin{table*}[t]
    \centering
    \caption{Retrieval results on LSMDC. * indicates that the method uses post-processing operations, ~\emph{e.g.,} DSL \cite{cheng2021improving} or QB-Norm \cite{bogolin2021cross}.}
    \resizebox{1\columnwidth}{!}{ 
    \begin{tabular}{l|ccccc|ccccc}
	\hline
	\multirow{2}{*}{Methods} & \multicolumn{5}{c|}{Text-Video}& \multicolumn{5}{c}{Video-Text}\\ 
    &R@1&R@5&R@10&MdR&MnR& R@1 & R@5 & R@10 & MdR & MnR\\ 
	\hline
	\hline
	CE\cite{liu2019use}&11.2&26.9&34.8&25.3&-&-&-&-&-&-\\
	CLIP4Clip\cite{luo2021clip4clip}&22.6&41.0&49.1&11.0&-&-&-&-&-&-\\
	CAMoE*\cite{cheng2021improving}&25.9&46.1&53.7&-&54.4&-&-&-&-&-\\
	MDMMT-2\cite{kunitsyn2022mdmmt}&26.9&46.7&55.9&6.7&48.0&-&-&-&-&-\\
	DCR*\cite{wang2022disentangled}&26.5&47.6&56.8&7.0&-&27.0&45.7&55.4&8.0&-\\
	\hline
	Ours$_{+B/16}$&24.5&44.1&53.9&9.0&52.5&24.6&43.6&51.8&9.0&47.8\\
    Ours$_{+L/14}$&27.1&45.1&53.4&8.5&60.3&26.1&42.5&52.0&9.0&54.5\\
	Ours$_{+L/14+DSL}$&$\boldsymbol{29.7}$&46.4&55.4&7.0&56.4&$\boldsymbol{30.1}$&47.5&55.7&7.0&48.9\\
    \hline
    Ours$_{huge}$&$\boldsymbol{40.4}$&80.1&92.8&2.0&3.9&$\boldsymbol{34.6}$&71.8&91.8&2.0&4.3\\
	\hline
    \end{tabular}}
    \label{tab:lsmdc}
    \vspace{-0.2cm}
\end{table*}

\begin{table*}[t]
    \centering
    \caption{Retrieval results on DiDeMo. * indicates that the method uses post-processing operations, ~\emph{e.g.,} DSL \cite{cheng2021improving} or QB-Norm \cite{bogolin2021cross}.}
    \resizebox{1\columnwidth}{!}{ 
    \begin{tabular}{l|ccccc|ccccc}
	\hline
	\multirow{2}{*}{Methods} & \multicolumn{5}{c|}{Text-Video}& \multicolumn{5}{c}{Video-Text}\\ 
    &R@1&R@5&R@10&MdR&MnR& R@1 & R@5 & R@10 & MdR & MnR\\ 
	\hline
	\hline
	CE\cite{liu2019use}&15.6&40.9&-&8.2&-&27.2&51.7&62.6&5.0&-\\
	ClipBERT\cite{lei2021less}&21.1&47.3&61.1&6.3&-&-&-&-&-&-\\
	FROZEN\cite{bain2021frozen}&31.0&59.8&72.4&3.0&-&-&-&-&-&-\\
	CLIP4Clip\cite{luo2021clip4clip}&41.4&58.2&79.1&2.0&-&42.8&69.8&79.0&2.0&-\\
	DCR*\cite{wang2022disentangled}&49.0&76.5&84.5&2.0&-&49.9&75.4&83.3&2.0&-\\
	\hline
	Ours$_{+B/16}$&45.0&75.6&83.4&2.0&12.0&45.7&73.3&83.8&2.0&8.6\\
	Ours$_{+L/14}$&41.8&71.2&79.0&2.0&17.2&39.7&71.7&79.8&2.0&12.9\\
	Ours$_{+B/16+DSL}$&$\boldsymbol{52.1}$&78.2&85.7&1.0&11.1&$\boldsymbol{54.8}$&79.9&87.2&1.0&7.1\\
	\hline
 Ours$_{huge}$&$\boldsymbol{52.7}$&77.8&85.2&1.0&13.7&$\boldsymbol{54.1}$&78.3&86.8&1.0&9.1\\
	\hline
    \end{tabular}}
    \label{tab:didemo}
    \vspace{-0.2cm}
\end{table*}

\begin{table*}[t]
    \centering
    \caption{Retrieval results on ActivityNet. * indicates that the method uses post-processing operations, ~\emph{e.g.,} DSL \cite{cheng2021improving} or QB-Norm \cite{bogolin2021cross}.}
    \resizebox{1\columnwidth}{!}{ 
    \begin{tabular}{l|ccccc|ccccc}
	\hline
	\multirow{2}{*}{Methods} & \multicolumn{5}{c|}{Text-Video}& \multicolumn{5}{c}{Video-Text}\\ 
    &R@1&R@5&R@10&MdR&MnR& R@1 & R@5 & R@10 & MdR & MnR\\ 
	\hline
	\hline
	CE\cite{liu2019use}&17.7&46.6&-&6.0&-&-&-&-&-&-\\
	SUPPORT\cite{patrick2020support}&28.7&60.8&-&2.0&-&-&-&-&-&-\\
	CLIP4Clip\cite{luo2021clip4clip}&41.4&73.7&85.3&2.0&-&-&-&-&-&-\\
	DCR*\cite{wang2022disentangled}&46.2&77.3&88.2&2.0&-&45.7&76.5&87.8&2.0&-\\
	\hline
	Ours$_{+B/16}$&46.4&79.2&90.0&2.0&4.6&45.4&78.7&89.9&2.0&4.8\\
	Ours$_{+L/14}$&44.1&77.3&88.5&2.0&5.2&41.6&74.4&87.4&2.0&5.5\\
	Ours$_{+B/16+DSL}$&$\boldsymbol{57.3}$&84.8&93.1&1.0&4.0&$\boldsymbol{57.7}$&85.7&93.9&1.0&3.4\\
	\hline
 Ours$_{huge}$&$\boldsymbol{59.2}$&86.1&93.4&1.0&3.7&$\boldsymbol{60.1}$&87.5&94.4&1.0&3.2\\
	\hline
    \end{tabular}}
    \label{tab:activity}
\end{table*}

Finally, $\lambda$ is evaluated in Table~\ref{fig:lambda}.
The results show that the marginal sample mining module obviously improves retrieval performance when $\lambda=0.1$.
When $\lambda>0.1$, the performance drops.
The reason may be that a strong constraint for hard samples hinders the model convergence at the training beginning.

\subsubsection{Visualization}
The core motivation of HCMI is to explore hierarchical cross-modal interactions between frame-word, clip-phrase, and video-sentence. 
Thus, we give some matching results between frames and words to verify the effectiveness of fine-grained interaction. 
From Figure~\ref{fig:vis}, most words can be matched with semantic-related frames.
Especially, some important nouns and verbs play important roles in TVR with large similarity scores, which justifies the effectiveness of exploring hierarchical semantic similarities.
Besides, we visualize auto-aggregated clip-phrase matching pairs in Figure~\ref{fig:vis_cp}.
For each clip and phrase, we only visualize top-3 tokens for simplicity.
From these results, HCMI can effectively aggregate semantic-related frames into an integrated clip and cluster important keywords into a phrase.
Since clips and phrases are learned from datasets, clip-phrase matching results are worse than frame-word matching.
Thus, this is one of our further research directions by exploring a stronger aggregation mechanism.

\subsection{Comparison with State-of-the-Art Methods}
In this part, we compare our HCMI with state-of-the-art methods on MSR-VTT, MSVD, LSMDC, DiDeMo, and ActivityNet benchmarks.

The MSR-VTT results are given in Table~\ref{tab:msrvtt}.
It can be seen that HCMI significantly surpasses CLIP4Clip by 10.4\% in R@1 and outperforms the brand-new method DCR by 1.6\% in R@1.
This is because that HCMI considers multi-level interactions between video and text modalities, which is ignored by CLIP4Clip and DCR.
It is noted that the reason for the weak performance of ViT-L/14 is that a large-scale model tends to overfit on a small-scale dataset.
Thus, for the large-scale LSMDC dataset in Tables~\ref{tab:msvd} and \ref{tab:lsmdc}, HCMI with ViT-L/14 obtains an obvious gain over the previous methods.
Besides, MDMMT-2 also adopts ViT-L/14 as its backbone, but HCMI has a stronger ability to capture cross-modal correlation, thereby yielding superior retrieval performance.

Then, we report the best results of HCMI with the DSL strategy. 
These results demonstrate that the large-scale model can benefit from cross-modal knowledge when having enough data for training.

Besides dense model, we further expand HCMI to huge model with about 17B parameters,
which is denoted as HCMI$_{huge}$ in the Tables. Especially on MSR-VTT and LSMDC with
large-scale videos, HCMI shows dominated superiority over dense models. This proves the
potential of large-scale models on tackling multi-modal tasks.

In summary, HCMI obtains new state-of-the-art performance on popular TVR benchmarks.

\subsection{Dicussion about application}
From the experimental results, HCMI can significantly improve text-video retrieval performance, thereby boosting industrial applications of this task.
For example, HCMI can be used in the Web\&App search and recommendation business to recommend relevant videos, when you give a paragraph of description.
The Internet advertising industry can also be beneficial from HCMI, because it can recommend the propse digital advertisement content for you, when you read a commodity article.
Besides, HCMI can be used to generate a paragraph of video description by searching a corresponding sentence for each video segment. The above potential applications show the importance of HCMI at both academia and industry.

\section{Conclusion}
In this paper, we revisited recent Text-Video Retrieval (TVR) methods and analyzed their pros and cons.
Considering comprehensive interaction between two modalities in human perception, we proposed a novel method, named Hierarchical Cross-Modal Interaction (HCMI), which hierarchically explores video-sentence, clip-phrase, and frame-word interactions to understand text-video contents.
Besides, two boosting strategies,~\emph{e.g.,} adaptive label denoising and marginal sample enhancement, were designed to further improve the performance.
Consequently, HCMI has been demonstrated to surpass the existing TVR methods on five benchmarks,~\emph{i.e.,} MSR-VTT, MSVD, LSMDC, DiDeMo, and ActivityNet, by a notable margin.

\bibliographystyle{splncs04}
\bibliography{ref}
\end{document}